\documentclass{nle}
\usepackage[utf8]{inputenc}
\usepackage[capitalize,nameinlink]{cleveref}
\usepackage{graphicx}
\usepackage{xcolor}
\usepackage{url}
\usepackage{CJKutf8}
\usepackage{amsmath,amssymb}
\usepackage{latexsym}
\usepackage{multirow}
\usepackage{bm}
\usepackage{tikz}
\usepackage{environ}
\usepackage{subfigure}

\title[Primal Meaning Recommendation via On-line Encyclopedia]{Primal Meaning Recommendation for Chinese Expressions via Descriptions in On-line Encyclopedia}
      
\author[Zhiyuan Zhang and others]
      {Z\ls H\ls I\ls Y\ls U\ls A\ls N\ns Z\ls H\ls A\ls N\ls G,\ns W\ls E\ls I\ns L\ls I,\ns J\ls I\ls N\ls G\ls J\ls I\ls N\ls G\ns X\ls U\\MOE Key Laboratory of Computational Linguistics, School of EECS, Peking University
      \and
      and
      \ns X\ls U\ns S\ls U\ls N\\MOE Key Laboratory of Computational Linguistics, School of EECS, Peking University\\
      Center for Data Science, Beijing Institute of Big Data Research, Peking University}

\received{02 Feb 2019; revised 12 Mar 2019}

\pagerange{\pageref{firstpage}--\pageref{lastpage}}
\pubyear{2019}

\usepackage{nle-chicago-patch}
\bibpunct{(}{)}{,}{a}{}{;}

\begin{document}

\label{firstpage}
\maketitle

\begin{abstract}
Polysemy is a very common phenomenon in modern languages. Under many circumstances, there exists a primal meaning for the expression. We define the primal meaning of an expression to be a frequently used sense of that expression from which its other frequent senses can be deduced. Many of the new appearing meanings of the expressions are either originated from a primal meaning, or are merely literal references to the original expression, e.g., apple (fruit), Apple (Inc), and Apple (movie).
When constructing a knowledge base from on-line encyclopedia data, it would be more efficient to be aware of the information about the importance of the senses.
In this paper, we would like to explore a way to automatically recommend the primal meaning of an expression based on the textual descriptions of the multiple senses of an expression from on-line encyclopedia websites. We propose a hybrid model that captures both the pattern of the description and the relationship between different descriptions with both weakly supervised and unsupervised models. The experiment results show that our method yields a good result with a P@1 (precision) score of 83.3 per cent, and a MAP (mean average precision) of 90.5 per cent, surpassing the UMFS-WE baseline by a big margin (P@1 is 61.1 per cent and MAP is 76.3 per cent).
\end{abstract}

\section{Introduction}
\label{sec:introduction}

Polysemy is a common phenomenon in many languages, where a word in different contexts has different meanings. Wordnet \citep{miller1995wordnet} and HowNet \citep{Dong2006Hownet} list different senses and their definitions of an expression. When a human thinks of an expression even without context, one usually has one prior understanding of the expression. For example, when hearing the word ``apple'', one is more likely to think of the fruit apple. From the point of word sense disambiguation (WSD) \citep{new_raganato2017word}, it would also be helpful if the model is aware of the importance of the senses. Most frequent sense (MFS) and sense distribution learning \citep{new_mccarthy2007unsupervised,new_bhingardive2015unsupervised,new_bennett2016lexsemtm,new_pasini2018two,new_mancini2016embedding} are two popular tasks to find the information about the importance of the senses. In this paper, we consider a variant task, the primal meaning. We define the primal meaning of an expression to be a frequent sense of that expression from which its other frequent senses can be deduced. Different from the origin of the expression, the primal meaning of an expression is a still frequently used sense nowadays and may not be the origin of the expression. Our definition is not lexicologically strict, but is more application oriented. A detailed definition will be given in Section~\ref{sec:task definition}.

In this paper, we want to explore the problem of recommending the primal meaning of an expression based on the textual descriptions of the expression in an on-line encyclopedia like Wikipedia\footnote{\url{https://en.wikipedia.org}} and Baidu encyclopedia.\footnote{\url{https://baike.baidu.com}}. Different from the set of search engines that aim to measure the relevance of the documents to the search expression, all the descriptions of an expression are about different aspects of the expression. Therefore, the task of primal meaning recommendation requires more than a literal match, but needs to understand the semantic meaning of the descriptions, and the relation between different senses (e.g., deduction relation). Take Baidu encyclopedia for example, for each expression, there can be multiple senses (see Table \ref{tab:intro example}). For each sense, there are both textual descriptions and images. The first paragraph of the textual descriptions can be seen as the summary of this sense.

There is also a problem in on-line encyclopedias where any people can edit the expressions. Many senses of an expression have very trivial meanings that people do not care about when searching for the definition or the meaning of an expression. For example, the second sense in the expression ``bright spot ''(\begin{CJK*}{UTF8}{gbsn}亮点\end{CJK*}) means the damage on the LCD screen, which is rarely referred to. When people search for the expression ``bright spot'' (\begin{CJK*}{UTF8}{gbsn}亮点\end{CJK*}), they are more likely to want to see the first meaning. The encyclopedia websites will be much user-friendly if the senses can be listed in order.

\begin{table}[ht]
\caption{An example expression "bright spot" in Baidu encyclopedia. Under each expression, there are multiple senses. For this expression, we list four senses. The first paragraph of different senses is used as the summary that represents the semantic meaning of this sense. }
\begin{minipage}{\textwidth}
\begin{tabular}{p{12cm}}
\hline\hline
\textbf{Expression}: Bright spot 
(\begin{CJK*}{UTF8}{gbsn}亮点 \end{CJK*})\\ \hline
\textbf{Sense 1}: Bright spot, Pinyin of which is liàng diǎn, is a metaphor of glorious and eye-catching people or things. It also refers to not completely plasticized particles on colored or opaque thermoplastic sheets, films, or molded articles, which appear as colorless, transparent spots when viewed against light.\\
(\begin{CJK*}{UTF8}{gbsn}亮点，拼音是liàng diǎn，指比喻有光彩而引入注目的人或事物。  也指有色或不透明的热塑料片材、薄膜或模制品上所含没有完全塑化的粒点，当其在对光观察对呈现为无色的透明斑点。\end{CJK*})\\  \hline
\textbf{Sense 2}: Bright spots are also known as bright spots on LCD screens, and are a physical impairment of LCD screens. This is mainly due to the fact that the internal reflector of the screen in the bright spot area is compressed by external forces or slightly deformed by heat.\\ 
(\begin{CJK*}{UTF8}{gbsn}亮点也被称为液晶显示屏亮斑，是一种液晶屏的一种物理损伤。主要是由于亮斑部位的屏幕内部反光板受到外力压迫或者受热产生轻微变形所致。\end{CJK*}) \\ \hline
\textbf{Sense 3}: Bright spot is a buzzword on-line, referring to the best, most funny, or most controversial part of a post.\\
(\begin{CJK*}{UTF8}{gbsn}亮点，网络流行语，指一个帖子中最精华、最搞笑或最引争议之处。\end{CJK*})\\ \hline
\textbf{Sense 4}: The particularly strong reflections appearing on the recorded section of the seismic exploration data after true amplitude recovery are called bright spots. The existence of bright spots may be a reflection of oil and gas reservoirs. Because the reflectivity of the gas-bearing formation interface is particularly large.\\
(\begin{CJK*}{UTF8}{gbsn}地震勘探资料经过真振幅恢复处理所得的记录剖面上出现的特别强的反射称为亮点。这种亮点的出现,可能是油气藏的反映。因为含气地层界面的反射系数特别大。\end{CJK*})\\ 
\hline\hline
\end{tabular}
\vspace{-2\baselineskip}
\end{minipage}
\label{tab:intro example}
\end{table}

Observed from the data, we find three characteristics that can help the model determine which the primal meaning is: 
\begin{enumerate} 
\item The description of the primal meaning is semantically more related to the literal meaning of the expression. 
\item Other senses of the expression can be deduced 
 from (derived from) the primal meaning, while the connection among other senses is small. For example, for the expression ``bright spot'' (\begin{CJK*}{UTF8}{gbsn}亮点\end{CJK*}) (see Table \ref{tab:intro example}), all the other senses are derived from the first one (the primal meaning), while the connection among the others is trivial.
 \item The expression pattern (style) of primal meaning is different. For example, for the expression ``bright spot'' (\begin{CJK*}{UTF8}{gbsn}亮点\end{CJK*}), the primal meaning (listed at the first) uses a lot of expressions about the definition, while other descriptions are more descriptive and less formal.
\end{enumerate}

Based on the observation above, we propose to use \textbf{Skip-Thought Model} (capturing the semantic similarity between the expression and description), \textbf{Relation Graph Model} (capturing the derivation relation between senses) and \textbf{Pattern Detection Model} (detecting the pattern or style of the descriptions) to capture the three characteristics of the primal meanings. Skip-Thought is an unsupervised model that can capture the semantic relation between sentences. Relation Graph and Pattern Detection are weakly supervised models. By weakly supervised, we mean the supervision signal is heuristically chosen and can be noisy \citep{Mintz2009}. 

In Baidu encyclopedia, definition files are like dictionaries where the document is clearly segmented into senses. Unlike dictionaries, the documents are created with crowdsourcing. The different senses are organized by the creating time. In front of an expression, Baidu encyclopedia gives all senses and their links so that one can click on a sense and get its description. The first created sense (listed in the first place) is more likely to be the primal meaning based on our observation. We use this characteristic for weakly supervised training. However, this is not absolute, which makes our weakly supervised training dataset noisy.

To train Pattern Detection and Relation Graph models, we propose to apply weakly supervised methods, which heuristically treat the first listed sense as the primal meaning in the weakly training set, while discarding the order information during the test stage. This is grounded on the fact that in about 44 per cent cases (1,479 out of 3,396) in a sample training data, the primal meaning is listed at the first place if there is a primal meaning.

To take advantage of the three models, we further propose to combine them together as one hybrid model. Experiment results show that our hybrid model is very effective. The results achieve a MAP score of 90.5 per cent and a P@1 score of 83.3 per cent.

Our contributions lie in the following aspects:
\begin{itemize}
\item We propose the task of recommending primal meaning of an expression in an on-line encyclopedia and manually annotate a dataset for evaluation.
\item Based on the observation of the data, we propose a hybrid model that uses both weakly supervised and unsupervised methods to solve the problem.
\item We do extensive experiments and make a detailed analysis of the advantages and weakness of our model. The experiment results show that our method is effective. 
\end{itemize}

\section{Methodology}
Based on the observation of the data and our understanding of the task, we design two kinds of models, unsupervised model (Skip-Thought model) and weakly supervised models (Pattern Detection and Relation Graph). Weakly supervised means that the supervision signal is heuristically set and noisy  \citep{Mintz2009}. The three proposed models are designed to capture three characteristics of the task. To take advantage of the three different aspects, we combine them together and get a hybrid model. In the following sections, we first describe the task definition, then we show the three models, finally, we show how to combine them together as the hybrid model.

\subsection{Task Definition}
\label{sec:task definition}

For an expression $e$ with $m$ senses. For each sense $s_i$, there is a textual description $d_i$ attached, consisting of $l_i$ characters. The task is to recommend the senses of the expression in order so that the primal meaning is listed in the front. 

Here we give our definition of primal meaning, among frequently used senses, if there is one sense from which all other senses can be deduced, we define this one as the primal meaning. Different from the origin of the expression, the primal meaning of an expression is a still frequently used sense nowadays and may not be the origin of the expression. Otherwise, if there is not one sense from which all other senses can be deduced or the deduction relation is not obvious, we define that the expression has no primal meaning. When we annotate the validation set and test set, we find only about 29 per cent (1,747 of 5,935) of expressions have a primal meaning, because many expressions are names shared by different people or locations.

\subsection{Unsupervised Model}

Assume that the primal meaning of an expression should be in the dominant position in the occurrences of an expression in the real-life text, which means that the embedding of the expression learned from the bulk real-life text should be most relevant to the primal meaning of the expression. 

Let $v_e$ be the vector of the expression $e$, $v_i$ be the vector of the description $d_i$. We assume that the closer the two vectors $v_e$ and $v_i$ are in the hyperspace, the more relevant $s_i$ and $e$ are, thus making $s_i$ more likely to be the primal meaning of the expression. We measure the similarity of $v_i$ and $v_w$ by the cos similarity of two vectors,
\begin{equation}
sim(s_i, e)=\cos(v_i^\text{T}, v_e) =  \frac{v_i^\text{T} v_e}{\|v_i\| \|v_e\|} 
\end{equation}
where $v_e$ is calculated as the average of the character embeddings of $e$. To get the $v_i$ of $d_i$ in an unsupervised way, we apply the Skip-Thought model \citep{Kiros2015Skip}. The intuition behind Skip-Thought is to predict the next and the last sentence based on the current one by sequence-to-sequence framework \citep{NIPS2014_5346} without attention mechanism, inspired by the skip-gram model \citep{Mikolov2013}. They assume that if the model can predict the next or last sentence, it must have some understanding of the current sentence. The sentence can then be represented as the last hidden state of the encoder, based on which the decoder predicts the next or the last sentence. We use the pretrained encoder in the Skip-Thought model to encode $d_i$ into the vector $v_i$.

\subsection{Weakly Supervised Model}

Observed from the training set,  we find that in about 44 per cent cases, the primal meanings are listed in the first place of the web page if there is a primal meaning of the sense. We use this characteristic to do weakly-supervised training by heuristically assuming that all the meanings listed in the first place in the training data are the primal meaning. By weakly supervised, we mean the supervision signal is heuristically set and noisy. We expect our model to recognize the pattern behind the noise. Based on the weakly supervised assumption, we treat the first sense, $s_1$ as the primal meaning. However, only about 29 per cent of senses have a primal meaning. For senses without a primal meaning, we also treat its first sense, $s_1$ as the primal meaning. Therefore, only a small portion of the supervision signal is exact. The supervision signal is weak and noisy.

\subsubsection{Pattern Detection Model}
\label{sec: pattern detection}

From the description data, we observe that the expression pattern of primal meanings themselves have some characteristics, e.g., the language style of primal meanings is usually more strict and precise. Therefore, we design a classifier that aims to recognize the pattern of the description of the primal meanings.

\begin{figure}[]
\centering
\includegraphics[width=3in]{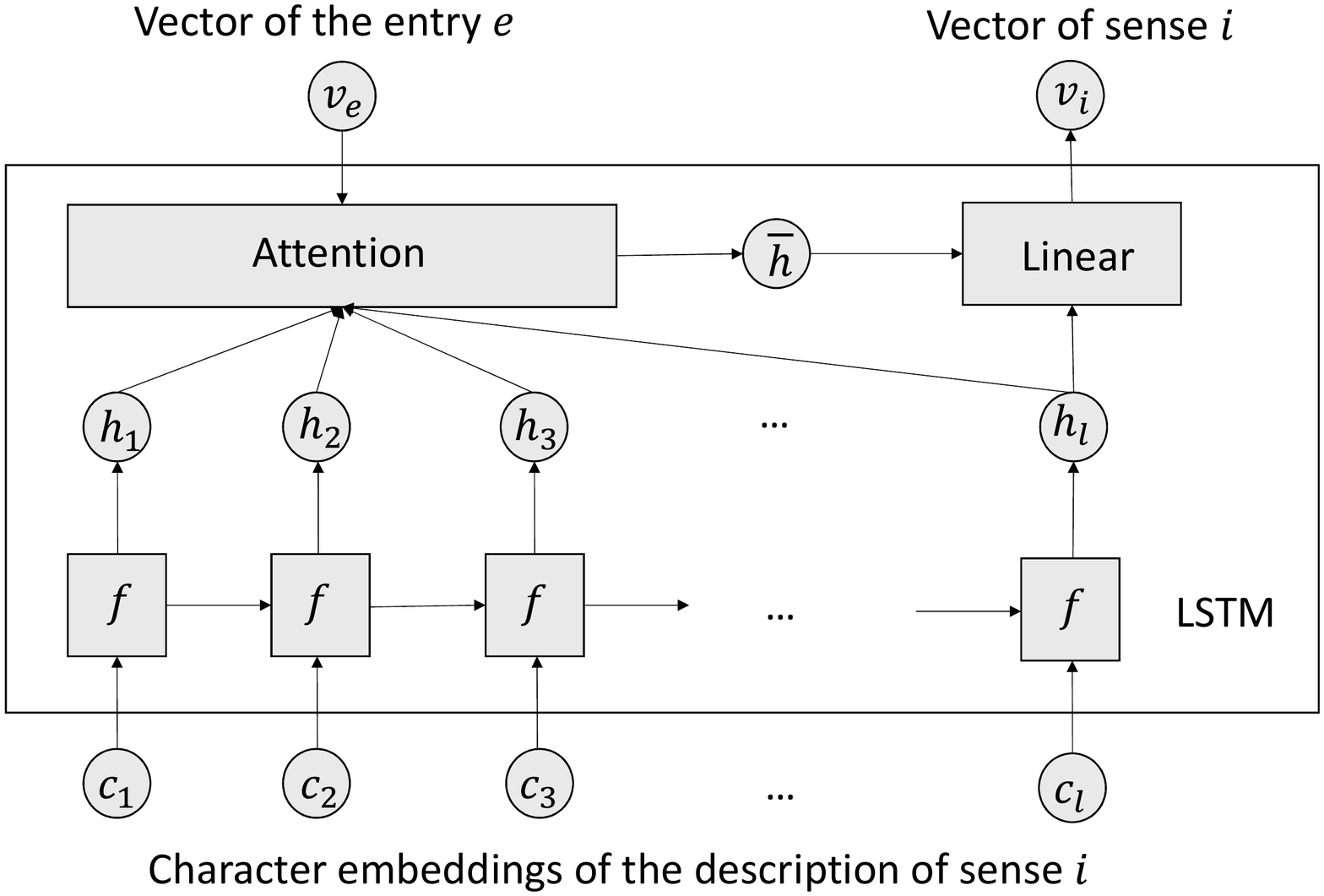}
\caption{An illustration of how to get the vector of sense $i$ with character embeddings of the description of sense $i$ the vector of expression. Here, $f$ denotes the recurrent non-linear function of LSTM cell, $v_i$ denotes the vector of sense $i$, $v_e$ denotes the vector of the expression $e$, $c_j$ denotes the character embeddings of the description and $h_j$ denotes the hidden state of LSTM.}
\label{fig:Pattern Detection}
\end{figure}

We apply Long Short Term Memory Networks (LSTM) \citep{RN1} to classify whether a sense of the expression is the primal meaning based on its description (see Figure \ref{fig:Pattern Detection}). For the description $d_i$ of sense $s_i$, assume there are $l_i$ characters $c_1, c_2, \cdots, c_{l_i}$. Each character $c_t$ is first projected to its character embedding $x_t$, then a recurrent non-linear function $f$ is applied to calculate the hidden state $h_t$, which gives us a sequence of hidden states $h_1, h_2, \cdots, h_{l_i}$,
\begin{equation}
h_t = f(h_{t-1}, E c_t)
\end{equation}
where $E\in R^{V\times d}$, $V$ is the vocabulary size and $d$ is the embedding dimension. Term $f$ is one step of LSTM.

Because the description $d_i$ can be very long, we use attention mechanism \citep{Wang2016Attention} to extract the useful information that is concerned about the expression $e$,
\begin{eqnarray}
v_i = \tanh(W_v [h_{l_i}; \bar{h}]) \\
\bar{h} = \sum_{j=1}^{l_i}\alpha_{j}h_{j} \\
\alpha_{j} = \frac{\text{exp}\left( a\left(v_e, h_{j}\right)\right)}{\sum\limits_{k=1}^{l_i}\text{exp}\left( a\left(v_e, h_{k}\right)\right)} 
\end{eqnarray}
where $v_e$ is the vector of expression $e$, which is calculated as the average of the character embeddings of expression $e$. Figure~\ref{fig:Pattern Detection} illustrates how to get $v_i$, the vector of sense $i$ with character embeddings of the description of sense $i$ and the $v_e$.

After we get $v_i$, the vector of sense $i$. We use a softmax layer to predict $p_i$, the probability that sense $i$ is the primal meaning of the expression,
\begin{equation}
p_i = \frac{\exp(w_p^\text{T} v_i)}{\sum\limits_{j=1}^{m}\exp(w_p^\text{T} v_j)}
\end{equation}

The loss function is the cross-entropy function.

\subsubsection{Relation Graph Model}
\label{sec: relation graph}

As is mentioned in Section~\ref{sec:introduction}, the primal meaning of an expression can have a deduction relationship with other meanings. Assume that the primal meaning of an expression has more connection with other meanings. Therefore, in this section, we model the relationship between the descriptions of the expression, which may give a hint on which one is the primal meaning.  

In this model, we also use weakly supervised signals that assume the sense listed at the first place is the primal meaning. We use the same network structure with different parameters to get the vector of senses.

We apply a score function $sim$ to measure the probability of deduction between two vectors $v_i$ and $v_j$. Note that the usage of bilinear function makes $sim(v_i, v_j)\neq sim(v_j, v_i)$, therefore, the relation graph of different senses is directed,
\begin{equation}
sim(v_i, v_j) = {v_i^\text{T} M v_j}
\end{equation}
where $M \in \mathbb{R}^{d\times d}$ and $d$ is the dimension of vectors $v_i$ and $v_j$.

According to the assumption the primal meaning of an expression has more connection with other meanings than the connection between other meanings, we calculate $\xi_i$ as the scores of sense $i$,
\begin{equation}
\xi_i = \frac{1}{m-1}\sum\limits_{j \ne i}sim(v_i, v_j)
\end{equation}

We apply a softmax function on $\xi_i$ to predict $p_i$, the probability that sense $i$ is the primal meaning of the expression,
\begin{equation}
p_i = \frac{\exp(\xi_i)}{\sum\limits_{j=1}^{m}\exp(\xi_j)}
\end{equation}

The loss function is the cross-entropy function.

\begin{figure}
\centering
\includegraphics[width=3in]{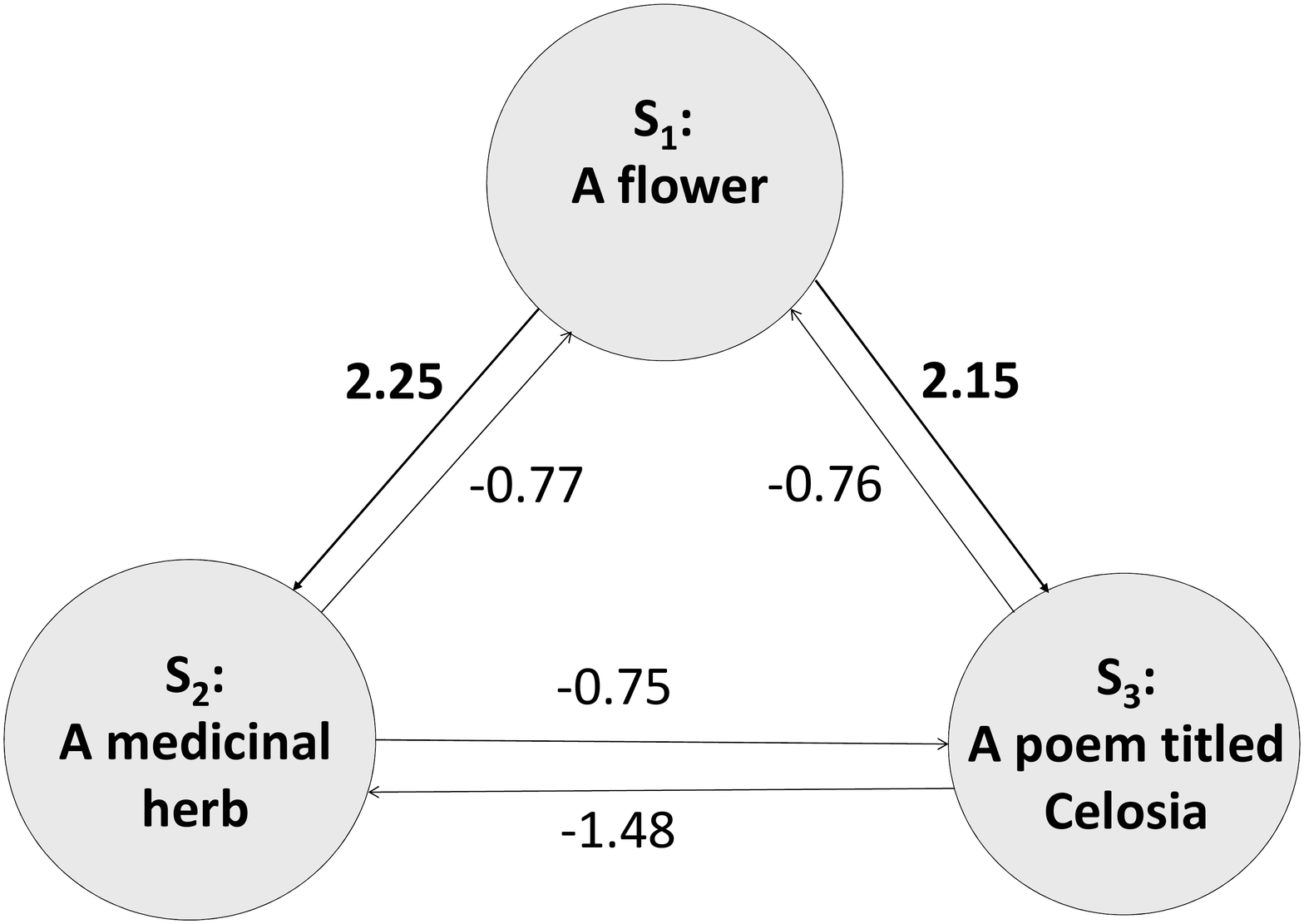}
\caption{An illustration of Relation Graph model. We use the expression ``Celosia'' as an example. The larger the edge is, the more likely there exists the deduction relation. The scores beside the edges are the real numbers calculated by our model. Bold arrows and numbers here denote the deduction relations.}
\label{fig:relation graph}
\end{figure}

If we view the description of a sense as a vertex, the deduction score $sim(v_i, v_j)$ as the edge, then we can get a directed relation graph that represents the deduction relation between senses. See Figure~\ref{fig:relation graph} for example. This is the relation graph of the expression ``Celosia'' (\begin{CJK*}{UTF8}{gbsn}鸡冠花\end{CJK*}), there are three senses for the expression. The primal meaning of the expression is $s_1$ (a flower).

\subsection{Hybrid Model} 
\label{sec:hybrid model}

Because different models capture different aspects of features, we combine the above three models, Pattern Detection, Relation Graph, and Skip-Thought together to form a hybrid model. 
Our hybrid model sums up the scores transformed from original scores given by the three models to be a new total score. The sense with the highest score is chosen as the primal meaning. Because the scales of the three models' scores can be different, we convert the original real number score into rank based score. The rank based score is calculated as follows,
\begin{eqnarray}
score_i^k = \frac{\text{exp}(-\lambda_k r_i)}{\sum\limits_{j=1}^{l_i}\text{exp}(-\lambda_k r_j)} \\
\lambda_k = \text{ln}(\frac{R_k}{R_k-1})
\end{eqnarray}
where $r_i$ is the rank of the score of a sense in the model $k$. $R_k$ is the average rank of the scores of the ground truth senses given by the model $k$. $\lambda_k$ is a hyper-parameter determined by $R_k$.

For example, if there are four senses in an expression, namely $(s_1, s_2, s_3, s_4)$. Assume the scores given by a model are $(0.4,0,1,0.2,0.3)$ and the mean rank result of the model is $2$, then $\lambda = \ln 2 = 0.693$. The ranks of the four senses are $(1, 4, 3, 2)$, their transformed score would be $(0.533, 0.067, 0.133, 0.267)$.

As we can see, the lower the mean rank result of model $k$ is, the higher the $\lambda_k$ is, namely its transformed scores will be more concentrated. If the model $k$ is perfect that the mean rank result approaches 1, then $\lambda_k$ approaches infinity, namely the primal meaning model $k$ predicts will get a transformed score approaching $1$, while others approaching $0$.
It can be mathematically proven that this simple transforming function can make the distribution reach the maximum entropy under some conditions.\footnote{The conditions and the proof are given in Appendix}

After we get the rank-based scores for each model, we calculate the total score by weighted sum the rank-based scores of the models. The weights are the P@1 of the individual models, which means the better the model performs individually, the more we trust its results,
\begin{equation}
total_i = \sum_{j=1}^3 p_j \times score_j ^ i
\end{equation}
where $total_i$ is the total score of the $i$-th sense, $3$ means the number of models (Pattern Detection, Relation Graph, Skip-Thought), $p_j$ is the P@1 performance of the $j$-th model on the development set. 

\section{Experiments}
\label{sec:experiment}
\subsection{Dataset Construction}

We crawl expressions with multiple meanings from Baidu encyclopedia.\footnote{\url{http://baike.baidu.com}} We use the summary of each sense provided by Baidu encyclopedia as the textual description of the sense. We also record the default order of different senses provided by Baidu, which is used in weakly supervised training but not available in the testing process.

We randomly choose a subset of the dataset and filter out the names of people and places, or other expressions that do not have the primal meaning by rule. Then we ask three human annotators to annotate the primal meaning of each expression. Before labeling the validation set and test set individually, the three persons label a small dataset commonly to ensure that they have the same understanding of the primal meaning of an expression. 
The agreement among three annotators is 92.1 per cent, that is, on 92.1 per cent of the cases, three annotators totally agree on what the primal meaning is. 

The crawled data other than the validation set and test set is used for weakly supervised training. In Table \ref{tab:dataset}, we show some statistic information about the dataset, including the number of all expressions (expressions), the average number of senses of all expressions (senses) and the average length of textual descriptions of senses (length). 

\begin{table}[ht]
\caption{Statistic information of the dataset}
\begin{minipage}{\textwidth}
\begin{tabular}{cccc}
\hline\hline
\bf Dataset & \bf Expressions & 
\bf Senses\footnote{The average number of senses of all expressions.} &
\bf Length\footnote{The average length of descriptions of the senses of all expressions.} \\ \hline
Noisy training set\footnote{Noisy training set for weakly supervision.} & 118,956 & 12\hpt 11 & 97\hpt 35 \\ 
Validation set\footnote{Human annotated validation set.} & 547 & 6\hpt 80 & 104\hpt 53 \\
Test set\footnote{Human annotated test set.} & 1,200 & 6\hpt 23 & 112\hpt 91 \\ 
\hline\hline
\end{tabular}
\vspace{-2\baselineskip}
\end{minipage}
\label{tab:dataset}
\end{table}

\begin{figure}[ht]
\centering
\begin{minipage}[ht]{0.5\textwidth}
\centering
\includegraphics[width=3.2in]{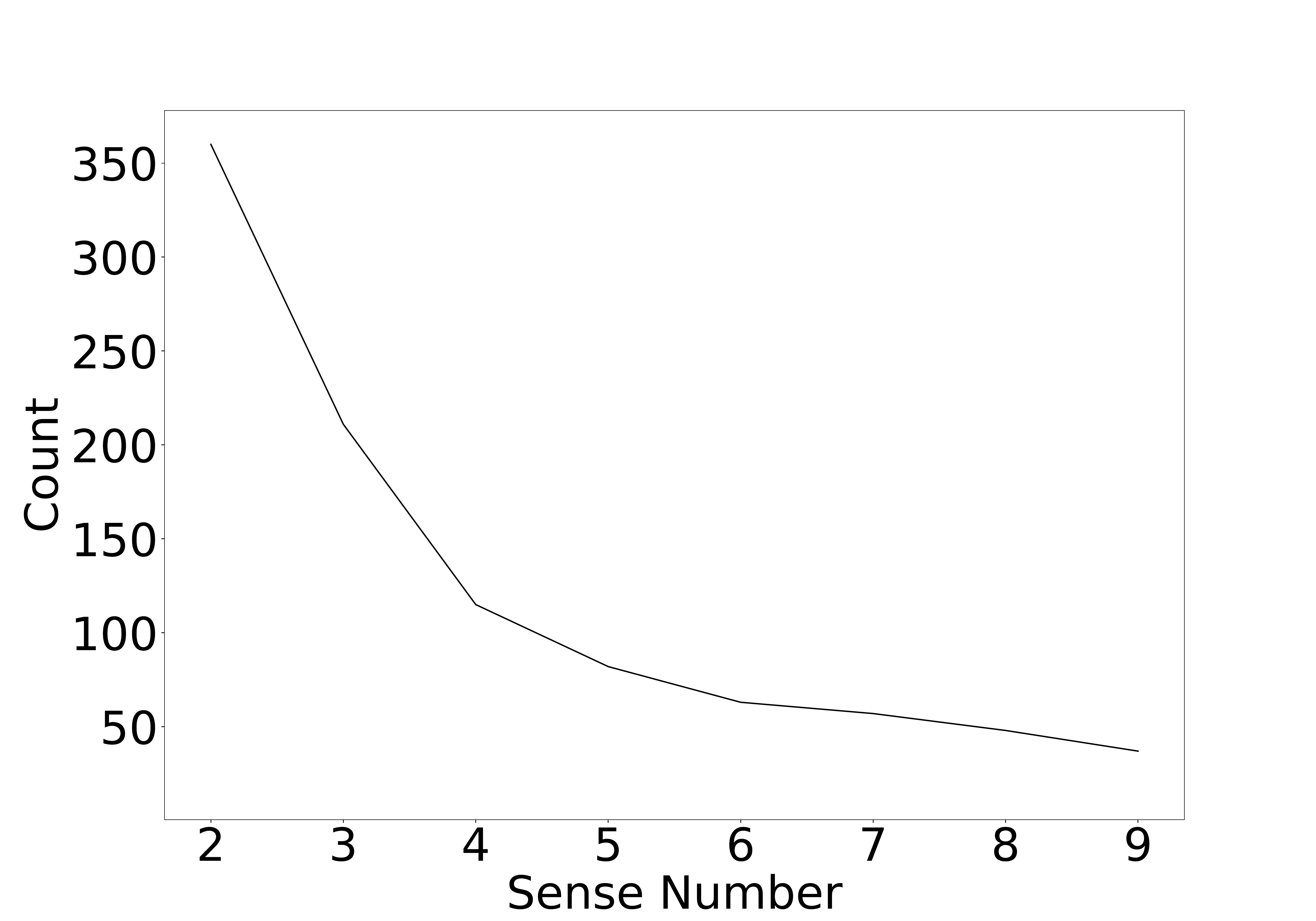}
\label{fig:Sense Number Distribution.}
\end{minipage}
\hfil
\begin{minipage}[ht]{0.5\textwidth}
\centering
\includegraphics[width=3.2in]{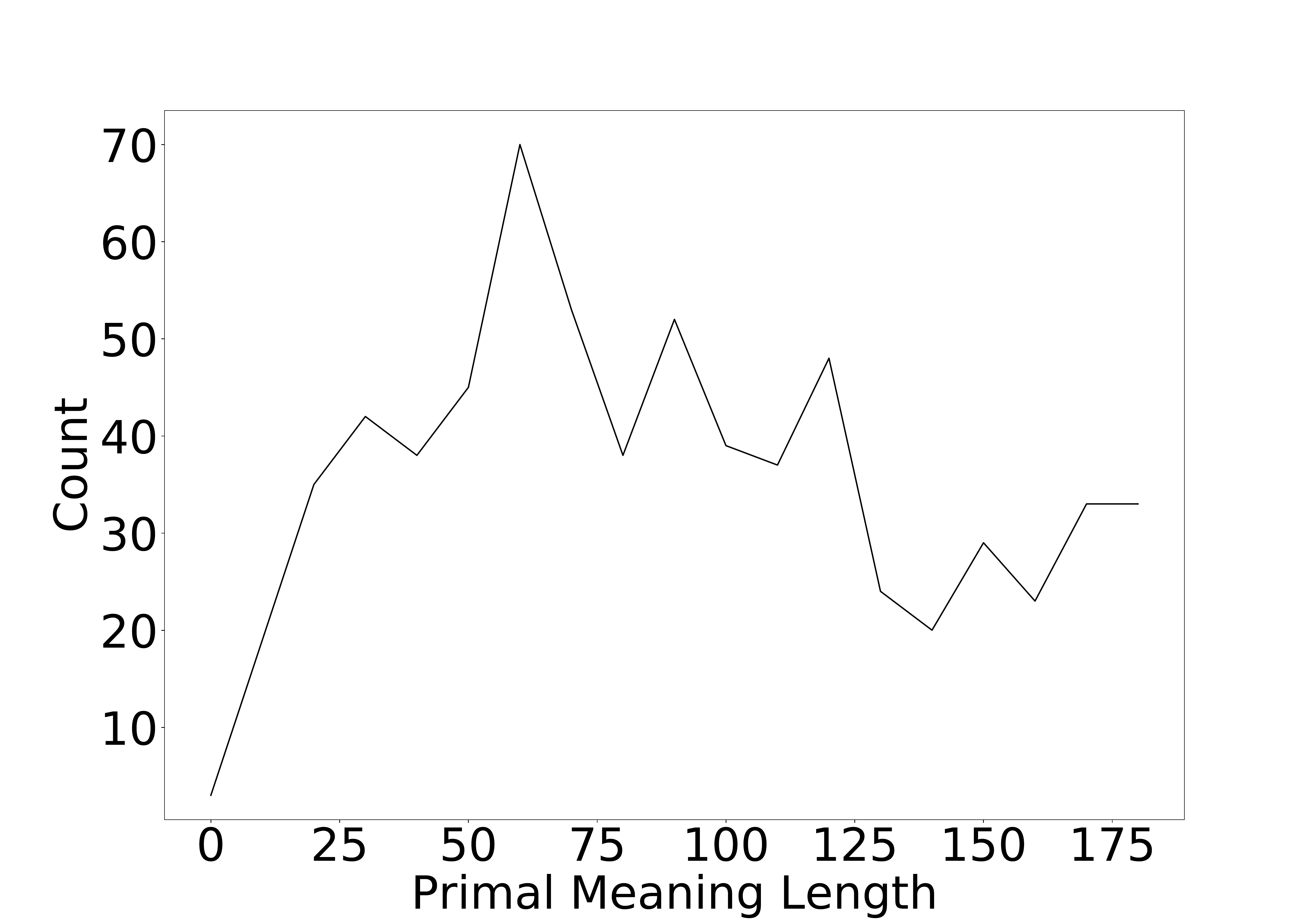}
\end{minipage}
\caption{Distribution of the number of senses for expressions (the first plot) and distribution of the length of descriptions (the second plot).}
\label{fig:Description Length Distribution.}
\end{figure}

The average number of senses in the validation set and test set are significantly smaller than that in the training set because many people or places share one name. The names of people and places usually have more senses than other expressions, which are filtered out in the validation set and test set. 

\subsection{Unsupervised Baseline}

\citet{new_bhingardive2015unsupervised} proposed \textbf{UMFS-WE} (Unsupervised MFS with word embeddings) baseline model for MFS. This model is similar to the Skip-Thought model where we compare the semantic similarity between $v_i$ and $v_e$. The difference is that instead of using a pretrained encoder with the Skip-Thought model, they use the average of the character embeddings of $d_i$ as $v_i$. Here we adopt UMFS-WE baseline for MFS as the baseline for our task.

\subsection{Experiment Details}

We use the character (instead of word) as a basic unit in our models because there are about 3,000 frequently used characters in Chinese. However, there are more than 120,000 frequently used words in Chinese. Different from Roman characters, Chinese characters maintain a large part of the semantic meaning. Furthermore, Chinese word segmentation will introduce new errors. 

The size of the character embedding is $300$. We set the size of the hidden state in LSTM as $200$. We use a two-layer bidirectional LSTM. The batch size is $8$. The dropout \citep{srivastava2014dropout} rate is $0.2$. The descriptions are truncated to a length of $200$. We use Adam algorithm \citep{Kingma2015} to optimize the model parameters. We train our models for $20$ epochs and use the parameters that gain the best accuracy score on the validation set.

\subsection{Results}

We use three metrics to evaluate our model, including P@1, mean average precision (MAP) \citep{Kishida2005Property} and mean rank. 

\begin{table}[ht]
\caption{Evaluation results of different models}
\begin{minipage}{\textwidth}
\begin{tabular}{lccc}
\hline\hline
\bf Model &  P@1 & MAP & Mean Rank  \\
\hline
   UMFS-WE & 61\hpt1 &  76\hpt3 & 1\hpt835  \\
\hline
   Skip-Thought & 64\hpt6 & 78\hpt8 & 1\hpt738  \\ 
   Pattern Detection & 79\hpt3 & 88\hpt0 & 1\hpt348  \\
   Relation Graph & 77\hpt0 & 86\hpt1 & 1\hpt463  \\
\hline
   Hybrid\footnote{Hybrid is the combination of Pattern Detection, Relation Graph and Skip-Thought.} & \textbf{83\hpt3} & \textbf{90\hpt5} & \textbf{1\hpt254}  \\ 
\hline\hline
\end{tabular}
\vspace{-2\baselineskip}
\end{minipage}
\label{tab:experiment result}
\end{table}

In Table \ref{tab:experiment result} we show the results of different models. The first two rows are the unsupervised models, the third and fourth row are the weakly supervised models.

Note that the hybrid model is the combination of Pattern Detection, Relation Graph, and Skip-Thought models, which is described in the hybrid model section. Weakly supervised models work better than unsupervised models. We assume that this is because our models can capture the cohesive pattern of the primal meanings in spite of the noise of the weak supervision signal. Unsupervised models can only make use of the semantic connection between the expression and the description of the sense, which may encounter some exceptions, and the semantic meaning is sometimes hard to get.

\begin{figure}[ht]
\centering
\includegraphics[width=3in]{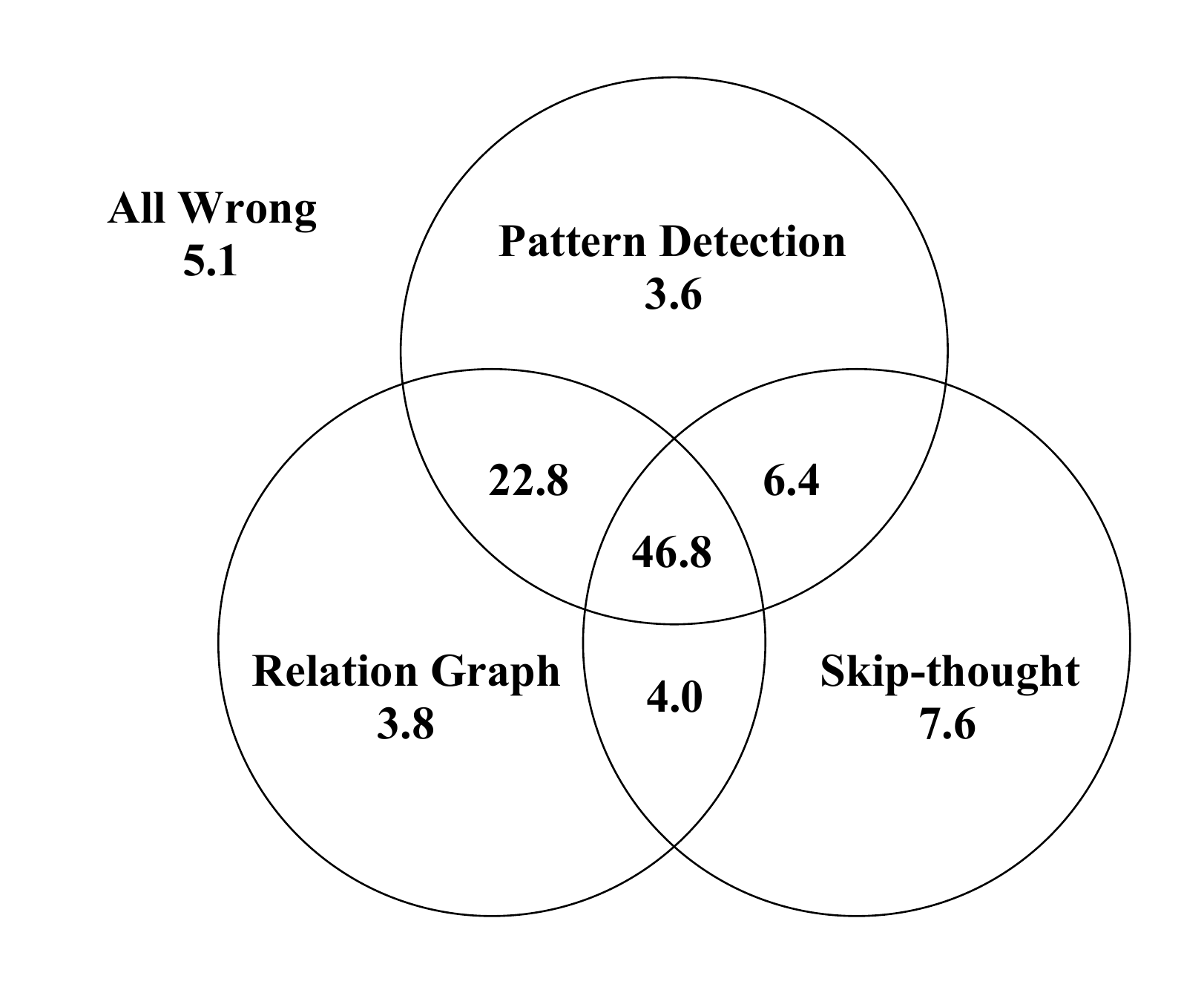}
\caption{Venn diagram of the P@1 results for different models. The overlap means both models are correct. The sum of percentages in different regions is not equal to 100 per cent because of rounding.}
\label{fig:Venn}
\end{figure}

We show the Venn diagram of P@1 results for different models in Figure \ref{fig:Venn}. From Figure \ref{fig:Venn}, we can see that the unsupervised model (Skip-Thought) is diverse from the weakly supervised models (Pattern Detection and Relation Graph), which means they can capture different aspects of primal meaning. However, the difference between the two weakly supervised models is not as big (17.8 per cent), we assume that this is because they are using the same weak supervision signal. Still, they are able to capture some different aspects because of the different design of their objective.

Among the individual models, Pattern Detection model works the best. We think that this is because the pattern of the descriptions of primal meanings is relatively clear compared with other senses. The expressions of many of the descriptions of the primal meanings are similar to that of dictionaries, while senses other than primal meanings often borrow the literal words as the name of either songs, movies or books. Actually, this is one of the situations where we need to recommend the primal meanings in the front because other senses are rare and trivial to most of the users.

The performance of Relation Graph model is close to Pattern Detection model, even though the working principle is different. Relation Graph can successfully capture the deduction relation between senses when the pattern of the descriptions is not obvious. Therefore, the two models are good complements to each other.

Skip-Thought model works better than the baseline UMFS-WE model, even though both the two models are unsupervised models because Skip-Thought model can better capture the semantic meaning of the sentences. The simple UMFS-WE model can be misled by literal similarity. From the experiment results, we observe that Skip-Thought model sometimes even beats weakly supervised models when both the pattern and the relation are not obvious and Skip-Thought model can capture the semantic meaning between the description and the expression itself, thanks to the big bulk of the pretraining data.

\begin{figure}[ht]
\centering
\begin{minipage}[ht]{0.5\textwidth}
\includegraphics[width=3.2in]{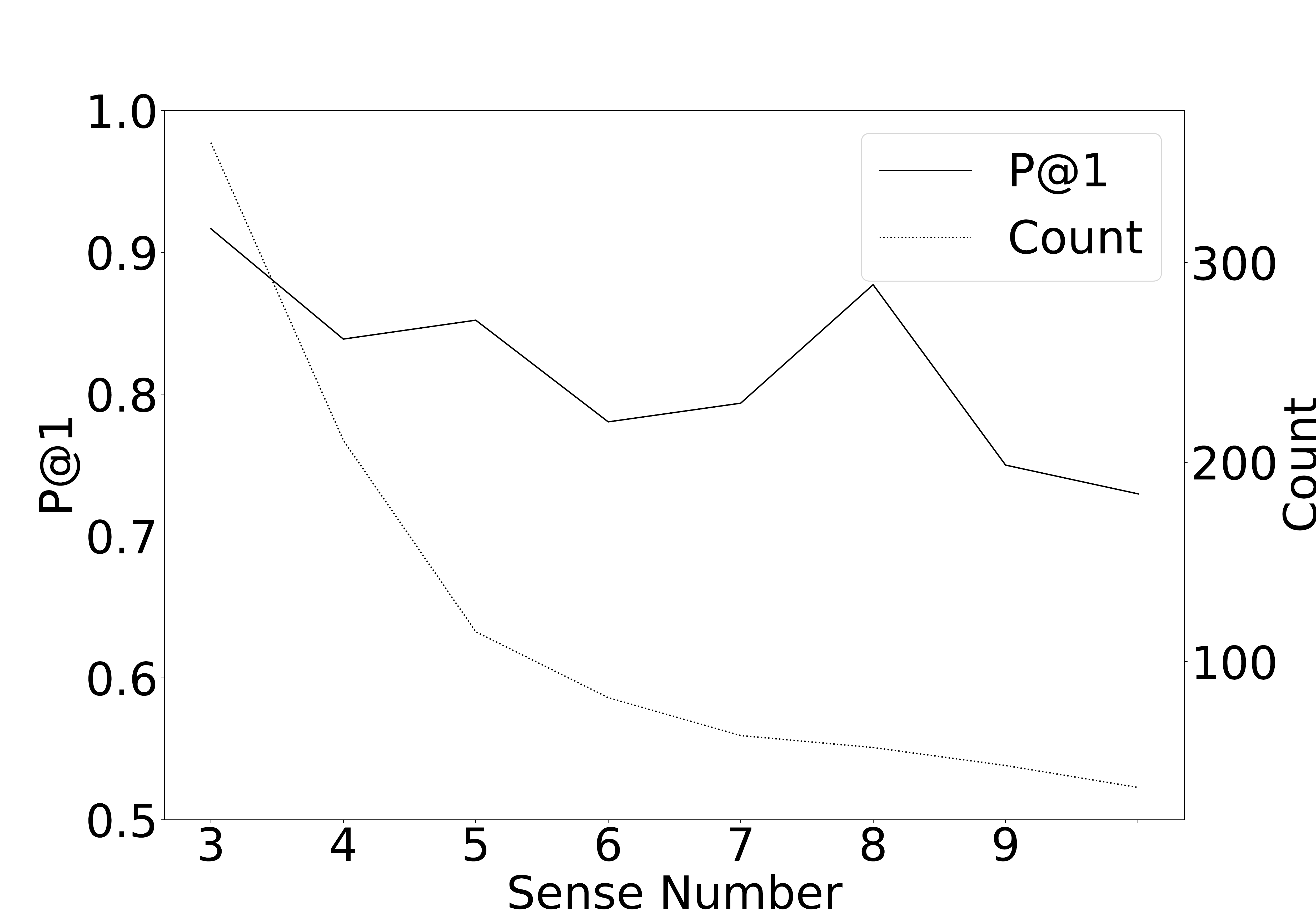}
\centering
\caption{Distribution of P@1 of the hybrid model along the number of senses in expressions.}
\label{fig:precision Sense Number Distribution.}
\end{minipage}
\end{figure}
\begin{figure}[ht]
\centering
\begin{minipage}[ht]{0.5\textwidth}
\centering
\includegraphics[width=3.2in]{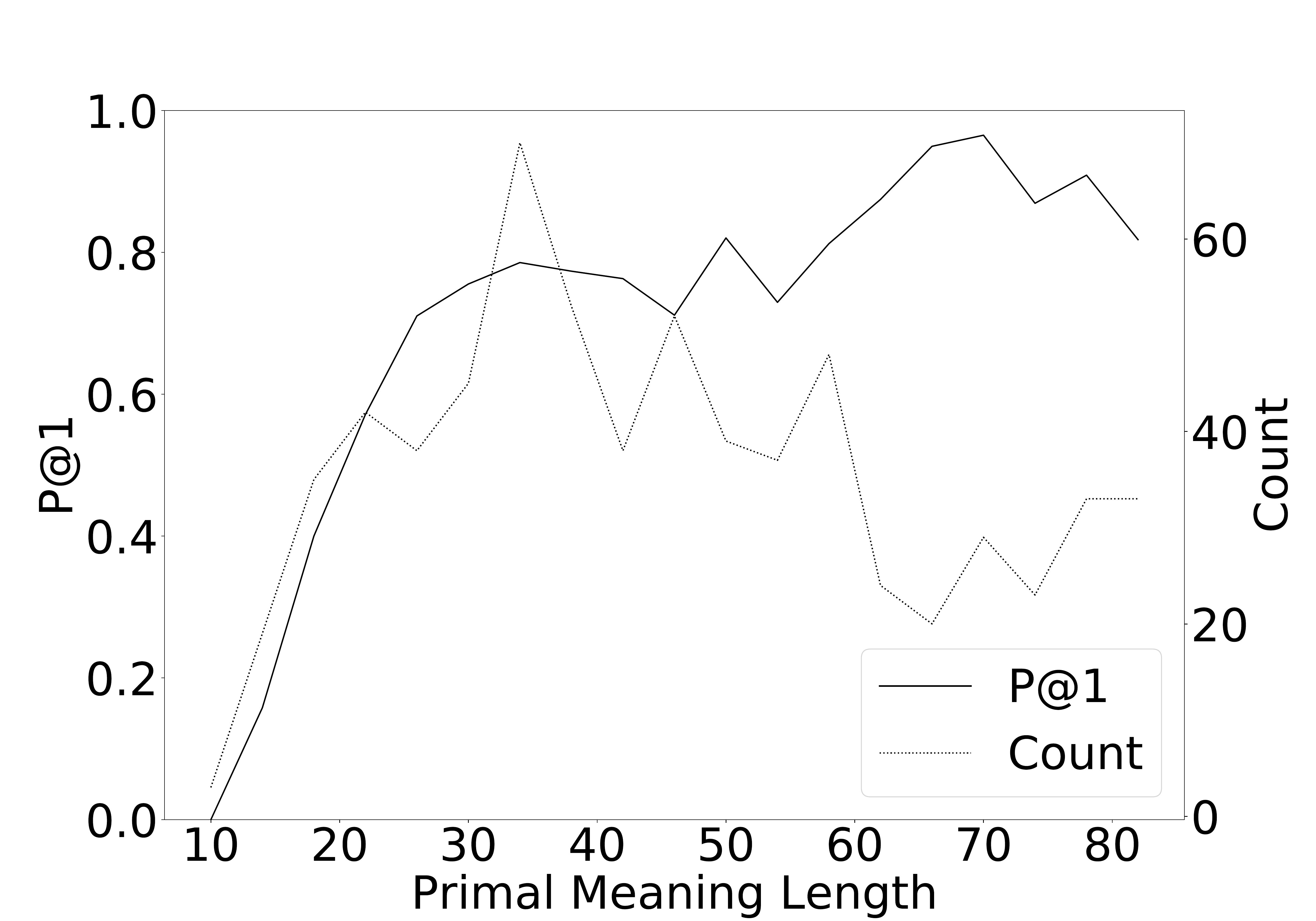}
\end{minipage}
\hfil
\begin{minipage}[ht]{0.5\textwidth}
\centering
\includegraphics[width=3.2in]{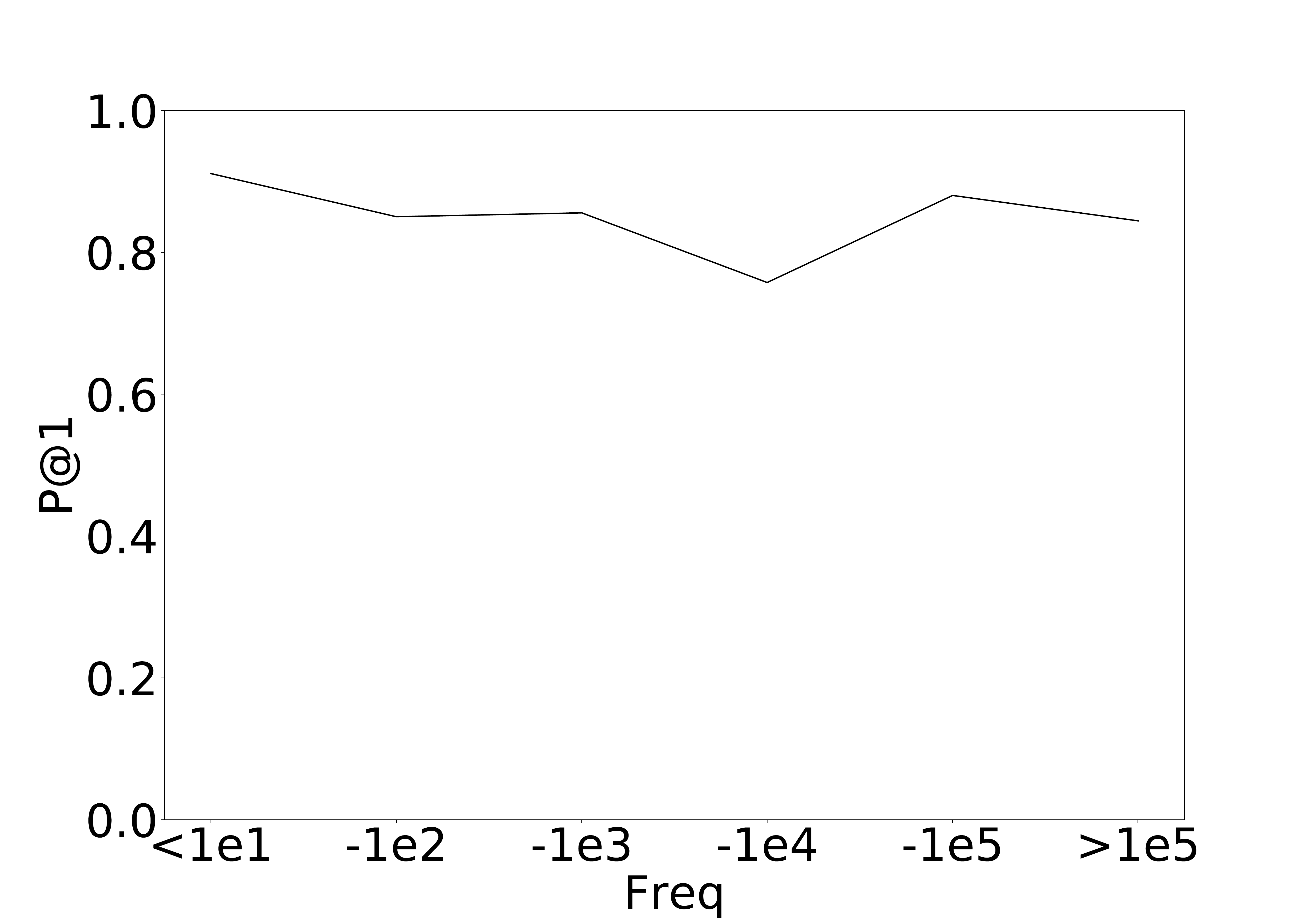}
\end{minipage}
\caption{Distribution of P@1 of the hybrid model along the length of primal meaning description (the first plot) and distribution of P@1 along the expression frequency in a big corpus (the second plot).}
\label{fig:Term Frequency Distribution.}
\end{figure}

From Figure \ref{fig:precision Sense Number Distribution.} we can see that the primary trend is that as the number of senses increases, P@1 decreases. This shows that the number of the senses in an expression can influence the accuracy (P@1), which is expected as the more senses there are the more difficult it is to find out what the primal meaning is. From the first part of Figure \ref{fig:Term Frequency Distribution.} we can see as the length of the primal meaning increases, P@1 increases. We assume that this is because the descriptions are more informative with longer text. From the second part of Figure \ref{fig:Term Frequency Distribution.} we can see that our model is not sensitive to the fluctuation of term frequency (TF), which shows that our model can work well for rare words, e.g., the expression ``gravity''.

\subsection{Error Analysis and Case Study} \label{sec:analysis}

In this section, we give some concrete examples of the advantages and weakness of various models.

\begin{table}[ht]
\caption{Example where UMFS-WE model makes a mistake. Sense number is the number of total senses of the  expression. The number of appearance (5 times) of the expression ``gravity'' affects the judgment of UMFS-WE model.}
\begin{minipage}{\textwidth}
\begin{tabular}{p{12cm}}
\hline\hline
\begin{CJK*}{UTF8}{gbsn} \textbf{Expression}:  (gravity) (万有引力 \end{CJK*})\\ \hline
\textbf{Sense Number}: 9 \\ \hline
{\textbf{Primal Meaning (the ground truth):}} The law of universal { \textbf{gravitation}} is a law that was discovered by Newton in 1687. There is a mutual attraction between any objects. The magnitude of this force is proportional to the mass of each object, and is inversely proportional to the square of the distance between them.\\ 
 (\begin{CJK*}{UTF8}{gbsn}{ \textbf{万有引力}}定律，物体间相互作用的一条定律，1687年为牛顿所发现。任何物体之间都有相互吸引力，这个力的大小与各个物体的质量成正比例，而与它们之间的距离的平方成反比
\end{CJK*})
\\\hline
{\textbf{Mistaken Sense (chosen by our model):}} ``{ \textbf{Gravity}}'' is a comics written by Michimi Murakami. His earlier work was the comic ``The { \textbf{Gravity}} Love Song HELP!'' and the Doujinshi ``The { \textbf{Gravitational}} Variations'' In 1996, he started the series of comic works serialized in the bi-monthly magazine Wyatt. The first part was completed. The second part, ``{ \textbf{Gravity}} EX'', was not serialized any more. On October 4, 2000, the TV animation ``{ \textbf{Gravity}}'' was broadcast.\\
 (\begin{CJK*}{UTF8}{gbsn}《\textbf{{  万有引力}}》是村上真纪创作的漫画，前作是漫画《\textbf{{  万有引力}}纯情曲 HELP!》以及同人志《\textbf{{  万有引力}}变奏曲》，于1996开始在幻冬社双月刊上开始连载的漫画作品，第一部已于完结。第二部《\textbf{{ 万有引力}}EX》处于休载状态。2000年10月4日，电视动画《\textbf{{ 万有引力}}》播出。\ \end{CJK*})\\
 \hline\hline
\end{tabular}
\vspace{-2\baselineskip}
\end{minipage}
\label{tab:error 1}
\end{table}

\noindent\textbf{Weakness of UMFS-WE Model: }In Table \ref{tab:error 1}, we show an example that UMFS-WE model fails to detect the primal meaning. The primal meaning of the expression ``gravity''  should be the physical law (the first meaning expression). However, because there are many occurrences of the expression ``gravity'' in the second sense listed in the table, the sum of the character embeddings becomes very close to the vector of the expression. This misleads the UMFS-WE model to take the description of the comics as the primal meaning. Other models make the right choice because they are better able to understand the semantic meaning of the description.

\begin{table}[ht]
\caption{Example where only Pattern Detection model works. The second sense is even more semantically relevant to the phrase expression than the primal meaning.}
\begin{minipage}{\textwidth}
\begin{tabular}{p{12cm}}
\hline\hline
\textbf{Expression}: Financial Regulations and Accounting Professional Ethics
(\begin{CJK*}{UTF8}{gbsn} 财经法规与会计职业道德\ \end{CJK*}) \\ \hline
\textbf{Sense Number}: 10 \\ \hline
\textbf{Primal Meaning (the ground truth)}: ``Financial Regulations and Accounting Professional Ethics'' is one of the subjects of accounting qualification examination. It is an essential subject of accounting qualification.\\ 
(\begin{CJK*}{UTF8}{gbsn}财经法规与会计职业道德是会计从业资格考试的考试科目之一。是会计从业资格的必考科目。\ \end{CJK*})\\ \hline
\textbf{Mistaken Sense (chosen by our model)}: 
``Financial regulations and accounting professional ethics'' is a book published by Machinery Industry Press in 2010, and the author is Zhu Dan. The book consists of five chapters, including \textbf{{ the accounting legal system, the payment and settlement legal system, the tax legal system, the fiscal legal system, and accounting professional ethics}}.\\
(\begin{CJK*}{UTF8}{gbsn} 《财经法规与会计职业道德》是2010年机械工业出版社出版的图书，作者是朱丹。本书共五章，包括\textbf{{ 会计法律制度、支付结算法律制度、税收法律制度、财政法律制度、会计职业道德}}等内容。\ \end{CJK*}) \\   \hline\hline
\end{tabular}
\vspace{-2\baselineskip}
\end{minipage}
\label{tab:error 2}
\end{table}

\noindent\textbf{Advantage of Pattern Detection Model:} For expression ``Financial Regulations and Accounting Professional Ethics'' (\begin{CJK*}{UTF8}{gbsn}财经法规与会计职业道德\end{CJK*}) (see Table \ref{tab:error 2}), the primal meaning refers to one of the subjects in the qualification exam of accounting. Other senses are all about the guideline books written for the exam published by different publishers. Because the senses of books dominate the explanation expressions,  which are semantically similar to each other, the Relation Graph model fails to recognize the sense about the exam as the primal meaning. Because the descriptions of the books also talk about concepts highly related to accounting, the UMFS-WE and Skip-Thought models fail in this case. However, since the pattern of the primal meaning here is typical, the Pattern Detection model successfully recognizes this one as the primal meaning.

\noindent\textbf{Advantage of Unsupervised Models:} For the expression ``magic'', the primal meaning should be about the entertaining act magic, both UMFS-WE and Skip-Thought model make the right choice because there are many expressions related to magic in the first (correct) description, such as performance, art, and curiosity. These words indicate that this description is highly related to the expression ``magic''. However, other models fail in this case, maybe this is because the description of the primal meaning is semantically complex, making the Pattern Detection model fail to capture the pattern. The senses other than the primal meaning are mostly names of songs or names of movies, so there is no obvious deduction relation, which makes the Relation Graph model fail in this case.

\noindent \textbf{Advantage of Relation Graph Model:} For the expression Celosia (\begin{CJK*}{UTF8}{gbsn}鸡冠花\end{CJK*}) (see Figure \ref{fig:relation graph}),  the primal meaning should be the flower ``Celosia''. Other senses are ``the medicine made from the flower'' and ``the poem named after the flower''. There exists strong derivation or deduction relation between the flower and the other two senses, but there is little connection between the medicine and the poem. Therefore, the Relation Graph model is very confident in this case and makes the right prediction. The actual relevant score of the graph in the experiment is shown in Figure \ref{fig:relation graph}. From the graph we can see that the relevant score actually shows the deduction relation between senses, $sim(v_1, v_2) = 2.25$, $sim(v_1, v_3) = 2.15$, these two scores indicate that the Relation Graph model thinks that senses $s_2$ and $s_3$ can be deduced from $s_1$, while scores in the other direction are very low ($-0.77$ and $-0.76$). This is because of the way we set our objective, which encourages the out-degree the primal meaning vertex to be larger and reduces the score on other edges in the graph.

\noindent\textbf{Difficult Case for All Models:} The phrase expression ``The Adventures of Pinocchio'' is a difficult case where all our models fail to recognize the primal meaning. The primal meaning of this expression should be the Italian fairy tale ``Pinocchio''. The other senses are all movies or TV shows derived from this fairy tale. The writing patterns are similar between descriptions because they are all related to literature works. What makes it difficult is that the semantic meaning of the content is similar as they all tell the same story in different versions or different forms. Under this circumstance, all the models fail to recognize the correct primal meaning. 

\section{Related Work}

Polysemy is a common phenomenon in many languages, where a word in different contexts has different meanings. Wordnet \citep{miller1995wordnet} and HowNet \citep{Dong2006Hownet} list different senses and their definitions of an expression. Many researchers focus on the task of word sense disambiguation (WSD) \citep{yarowsky1992word,yarowsky1995unsupervised,banerjee2002adapted,WordClustering,Camacho-Collados2015}, which aimed to map an ambiguous word in a given context to its correct meaning. Most frequent sense (MFS) and sense distribution learning \citep{new_mccarthy2007unsupervised,new_bhingardive2015unsupervised,new_bennett2016lexsemtm,new_pasini2018two,new_mancini2016embedding} are also two popular tasks to find the information about the importance of the senses. \citet{new_bhingardive2015unsupervised} proposed an unsupervised MFS detecting method using word embeddings. \citet{new_mccarthy2007unsupervised, new_mancini2016embedding} proposed knowledge-based MFS detecting methods, which, however, are not available in our task because expressions in Baidu encyclopedia are in wide fields and Chinese version of Wordnet or Hownet cannot cover them.

Crawling data from on-line encyclopedias or knowledge
bases is an important way to construct the training data. \citet{LiWeiArxiv} crawled the data from TCM Prescription Knowledge Base for training an end-to-end method to generate traditional Chinese medicine prescriptions. \citet{LiuShuArxiv} made use of knowledge base to evaluate semantic rationality of a sentence. Recommender systems are designed to retrieve relevant content of query \citep{Ando2005,ricci2011introduction}. Our task can be seen as recommending the most important (primary) sense of words. Our task is also related to the question answering problem \citep{Berant2013,Yang2015,Qiu2015,QuestionConsiderQA,semi-structQA}. However, in the task of primal meaning recommendation, the semantic meanings are often very similar, there are usually many overlaps of words between senses. Therefore, a traditional recommender system or QA system is not suitable for this task.

\section{Conclusion}
In this paper, we propose the task of primal meaning recommendation based on the descriptions of on-line encyclopedia websites. Our work is to explore a way to organize senses of an expression with different importance. We propose to apply Skip-Thought, Pattern Detection, Relation Graph and a hybrid model to deal with the problem. We use the real-life on-line encyclopedia website Baidu encyclopedia to train and test our model. The final hybrid model achieves very good result on our human annotated test set with a P@1 score of 83.3 per cent and MAP of 90.5 per cent, surpassing the UMFS-WE baseline by a big margin (P@1 61.1 per cent  and MAP 76.3 per cent), which means our method can indeed help recommend the primal meaning of a word in the front. 

This work focuses on Chinese on-line encyclopedia. However, this work is not tied to Chinese and can be extended to other languages. When extended to another language, we can use word-based or character-based LSTM models. Only an on-line encyclopedia with weak supervision signals in this language is needed. In the future, we would like to explore the application of this work in downstream tasks like WSD, knowledge base construction and entity linking.

\section{Acknowledgements}
We thank the anonymous reviewers for their thoughtful comments. This work was supported in part by the National Nature Science Foundation of China (NSFC) under Grant 61673028. Xu Sun is the corresponding author of this paper.

\appendix

\section{Supplemental Material}
\label{sec:supplemental}

\subsection{Puzzle Definition}

In Section~\ref{sec:hybrid model}, we claim that it can be mathematically proven that our rank-based transformation function can make the distribution reach the maximum entropy under two conditions:
\begin{itemize}
\item The distribution sums to 1.
\item The expectation of the rank of the primal sense equals to the mean rank result of the model given our transformed score as the distribution.
\end{itemize}

In this section, we give the proof of our conclusion.

Suppose $p_i$ is the transformed score of the $i$-th original score of a single model in descending order, $n$ is the number of senses of this expression and $R$ is the mean rank score of this single model.

We want to find the maximum entropy under two conditions:
\begin{equation}
\begin{aligned}
\text{max} H(p_i)&=\sum_{i=1}^n{p_i\ln(p_i)}\\
s.t.\ \sum_{i=1}^n{p_i}=1 \quad &\text{and} \quad
\sum_{i=1}^n{ip_i}=R
\end{aligned}
\end{equation}

\subsection{Solution to the puzzle}

We imply lagrangian multiplier method,
\begin{equation}
L(p_i)=\sum_{i=1}^n{p_i\text{ln}(p_i)}-\alpha(\sum_{i=1}^n{p_i}-1)-\beta(\sum_{i=1}^n{ip_i}-R)
\end{equation}

When $p_i$ reaches the the maximum entropy,
\begin{equation}
\frac{\partial{L(p_i)}}{\partial{p_i}}=-1-\ln(p_i)+\alpha+\beta i=0
\end{equation}
which gives, 
\begin{equation}
p_i=\exp(\beta i +\alpha-1)=C\exp(-\lambda i) 
\end{equation}

According to the first condition,
\begin{equation}
\sum_{i=1}^n{p_i}=C\sum_{i=1}^n{\exp(-\lambda i)}=1
\end{equation}
which gives,
\begin{equation}
C=(\sum_{i=1}^n{\exp(-\lambda i)})^{-1}
\end{equation}

Here we solve the rank-based transformation function that can reach the maximum entropy under the two conditions.

\subsection{Choice of hyper-parameter $\lambda$}

In this section, we give the solution to the choice of hyper-parameter $\lambda$. 

According to the second condition,
\begin{equation}
\sum_{i=1}^n{i p_i}=C\sum_{i=1}^n{i\exp(-\lambda i)}=R
\end{equation}
therefore,
\begin{equation}
R\sum_{i=1}^n{\exp(-\lambda i)}=\sum_{i=1}^n{i\exp(-\lambda i)}
\end{equation}
after simplification,
\begin{equation}
\frac{n\ {\exp(-\lambda)}^{n+1}(1-\exp(-\lambda))}{\exp(-\lambda)-{\exp(-\lambda)}^{n+1}}=1-(1-\exp(-\lambda))\ R
\end{equation}

The average of $n$ is about $7$ in our validation set. Assume $n\exp(-\lambda)^{n+1}<<1$, the left hand side can be omitted,
\begin{equation}
0=1-(1-\exp(-\lambda))R
\end{equation}
we get,
\begin{equation}
\lambda=\ln(\frac{R}{R-1})
\end{equation}

We will check our assumption now. Assume $\lambda$ is about $1.2$ and $n$ is about $7$, then,
\begin{equation}
\text{left hand side}=\frac{n{\exp(-\lambda)}^{n+1}(1-\exp(-\lambda))}{\exp(-\lambda)-{\exp(-\lambda)}^{n+1}} = 0.00011
\end{equation}
which is very small and can be omitted, so our assumption is reasonable.

\bibliographystyle{chicago-nle}
\bibliography{bibliography}

\label{lastpage}
\end{document}